\documentclass[letterpaper, 10 pt, conference]{ieeeconf}  

\IEEEoverridecommandlockouts                              

\let\labelindent\relax

\usepackage{graphics} 
\usepackage{epsfig} 
\usepackage{mathptmx} 
\usepackage{times} 
\usepackage{amsmath} 
\usepackage{amssymb}  
\usepackage{graphicx}
\usepackage{tabularx}
\usepackage{multicol}
\usepackage[bookmarks=true]{hyperref}
\usepackage{xcolor}
\usepackage[ruled,vlined, linesnumbered]{algorithm2e} \SetArgSty{textup} \usepackage[noend]{algpseudocode}
\usepackage{subcaption}
\usepackage[noadjust]{cite}
\usepackage{adjustbox}
\usepackage{comment}
\usepackage[none]{hyphenat}
\usepackage{textcomp}

\DeclareCaptionFont{8pt}{\fontsize{8}{8.4}\selectfont}
\usepackage{multirow}
\usepackage{titlesec}
\titlespacing*{\section}{0pt}{0.37\baselineskip}{0.1\baselineskip}
\titlespacing*{\subsection}{0pt}{0.38\baselineskip}{0.2\baselineskip}

\usepackage{enumitem}
\setlist{nosep} 

\usepackage{microtype}
\renewcommand{\paragraph}[1]{\noindent {\bf #1}}

\renewcommand{\baselinestretch}{0.97}
\renewcommand{\baselinestretch}{0.982}

\usepackage[frozencache=true,cachedir=minted-cache]{minted}

\title{\LARGE \bf
CushSense: Soft, Stretchable, and Comfortable Tactile-Sensing Skin \\for Physical Human-Robot Interaction
}
\author{Boxin Xu$^{*}$, Luoyan Zhong$^{*}$, Grace Zhang, Xiaoyu Liang, Diego Virtue, \\Rishabh Madan, Tapomayukh Bhattacharjee
\thanks{* denotes equal authorship.}
\thanks{This work was done when all the authors were affiliated with Cornell University, Ithaca, NY, USA,
        {\tt\small \{bx65, lz572, gtz4, xl434, dtv25, rm773, tb557\}@cornell.edu}}
}

\begin{document}

\maketitle
\thispagestyle{empty}
\pagestyle{empty}

\begin{abstract}

Whole-arm tactile feedback is crucial for robots to ensure safe physical interaction with their surroundings. This paper introduces CushSense, a fabric-based soft and stretchable tactile-sensing skin designed for physical human-robot interaction (pHRI) tasks such as robotic caregiving. Using stretchable fabric and hyper-elastic polymer, CushSense identifies contacts by monitoring capacitive changes due to skin deformation. CushSense is cost-effective ($\sim$US\$7 per taxel) and easy to fabricate. We detail the sensor design and fabrication process and perform characterization, highlighting its high sensing accuracy (relative error of 0.58\%) and durability (0.054\% accuracy drop after 1000 interactions). We also present a user study underscoring its perceived safety and comfort for the assistive task of limb manipulation. We open source all sensor-related resources on \href{https://emprise.cs.cornell.edu/cushsense}{\textbf{emprise.cs.cornell.edu/cushsense}}.
\end{abstract}
\section{Introduction}

According to a 2021 report \cite{abdi2021emerging}, over a billion individuals globally require assistance with activities of daily living (ADLs), such as dressing and transferring. While caregiving robots hold the potential to assist these individuals, a significant challenge persists: existing robots cannot perform these tasks without making whole-arm contact, posing risks in safe pHRI \cite{grice2013whole}. A promising approach is using a robot arm with a whole-arm tactile-sensing skin, which can provide detailed information about contacts and promote safer physical interactions. Moreover, these sensors are not limited to caregiving; they are valuable when operating in unstructured environments where whole-arm contacts are inevitable, for instance, when retrieving objects from cluttered spaces \cite{jain2013reaching}. An ideal tactile-sensing skin for pHRI should be soft, offering active compliance through compliant control \cite{chengskincontrol} and passive (hardware) compliance for increased comfort. The skin should be stretchable, allowing coverage around the robot joints \cite{bhattacharjee2013tactile}, even during changes in the robot configuration. In addition, a well-designed tactile-sensing skin must exhibit high accuracy, repeatability, and durability for reliable force measurements and sustained use. It should also provide good spatial and temporal resolution for fine contact-rich manipulation. Lastly, the skin should be modular and cost-effective, ensuring easy access and use across a broad spectrum of robotic applications.

While rigid whole-arm tactile-sensing skins \cite{cheng2019comprehensive, zhou2023tacsuit} have shown efficacy in manipulating large objects and detecting obstacles in unstructured environments, they fall short in softness and stretchability. Building sensors that combine these qualities without compromising sensing capabilities is challenging. Fabric-based skin sensors \cite{bhattacharjee2013tactile, si2023robotsweater} offer some promise, but existing designs lack the requisite softness and cushioning for pHRI. Also, the large-scale nature of these skins introduces additional complexities. Due to their extensive coverage area and potential for multiple interaction points, these sensors are inherently more prone to noise from environmental interference, crosstalk, etc. The scale of these sensors also presents challenges in efficient fabrication and user-friendly assembly.

\begin{figure}[!t]
    \centering
    \includegraphics[width=0.42\textwidth]{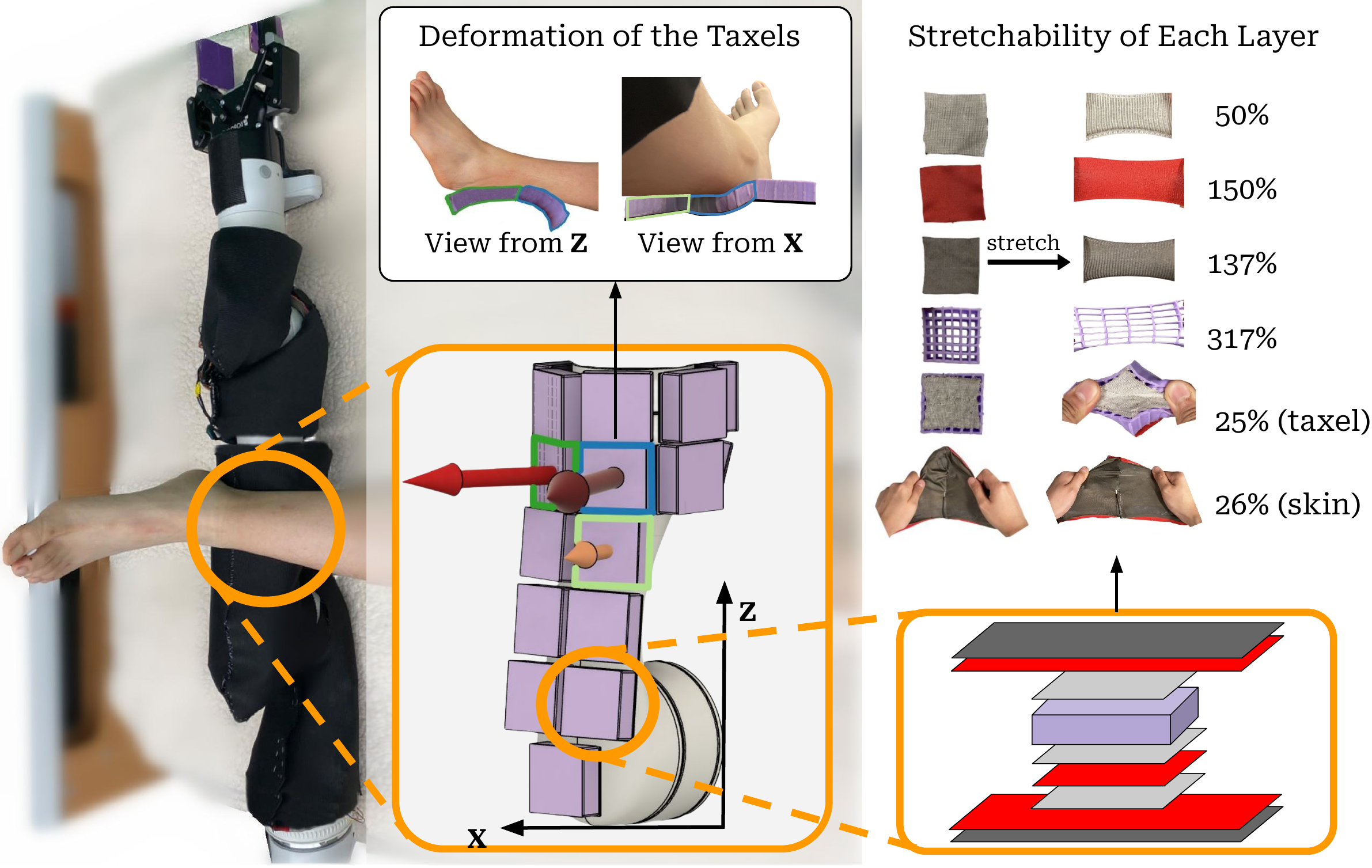}
    \caption{\textbf{Whole-Arm Soft and Stretchable Robot Skin.} We cover a robot arm with CushSense to perform limb manipulation. CushSense comprises taxels which use soft and stretchable materials, allowing each layer to stretch while maintaining reliable sensing and comfort.}
    \vspace{-15pt}
    \label{fig:wow_figure}
\end{figure}    

\begin{figure*}[ht!]
    \vspace{5pt}
    \centering
    \includegraphics[width=0.93\linewidth]{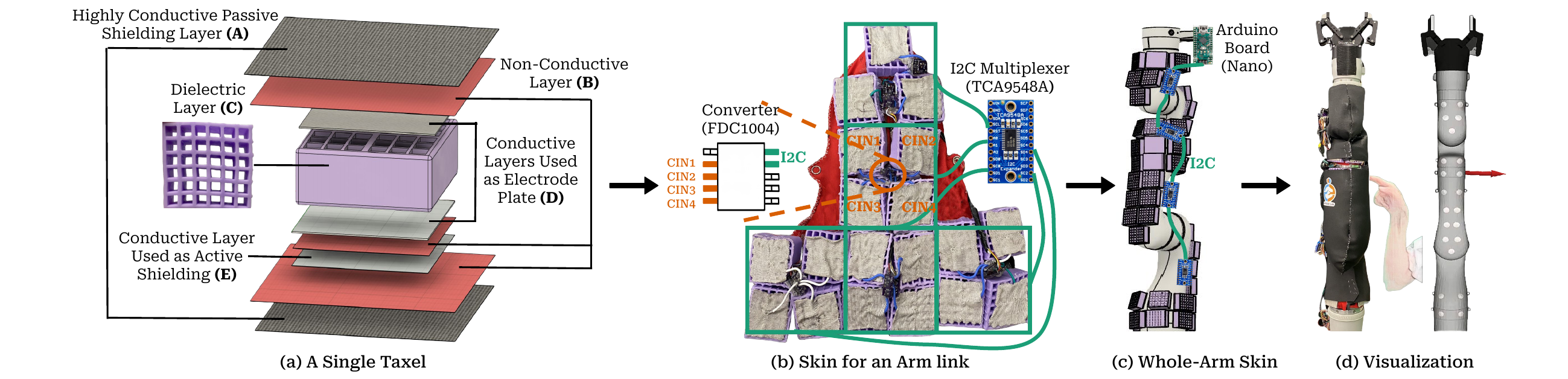}
    \caption{\textbf{Whole-arm Skin Structure}. (a) A taxel with 9 layers acts as a capacitive sensor. (b) An array of taxels forms a skin section, where each taxel measures a capacitance value. (c) We daisy-chain these sections to build a whole-arm skin comprising 56 taxels. (d) An Arduino board collects data and publishes it on a ROS topic, used to estimate contact forces, as shown on a 3D robot model in RViz.}
    \label{fabrication process}
    \vspace{-8pt}
\end{figure*}

We present \textit{CushSense} (see Fig. \ref{fig:wow_figure}), a fabric-based tactile-sensing skin that is soft, stretchable, and comfortable. CushSense comprises taxels (\textbf{tac}tile pix\textbf{el}s) made from low-cost stretchable fabric, a hyper-elastic polymer that provides the cushioning effect necessary for passive compliance, making the interaction more comfortable. It can detect taxel deformations resulting from applied forces by measuring changes in capacitance. With calibration, it can estimate normal forces and identify contact poses as the centers of activated taxels, making it useful for future applications that rely on tactile feedback for closed-loop control. To make the skin accessible, we used cost-effective materials ($\sim$US\$7 per taxel) and open-source fabrication methods that do not require access to specialized equipment.

The main contributions of this paper include:
\begin{itemize}
    \item Design and fabrication of a whole-arm soft and stretchable tactile-sensing skin.
    \item Sensor characterization demonstrating the sensing advantages of the proposed sensor design.
    \item pHRI user study highlighting the superior perceived safety and comfort of our sensor. 
    \item Open-source hardware release, including design files, fabrication process, calibration, and visualization utilities for the skin.
\end{itemize}
CushSense provides a convenient way of making rigid robots safer without significantly sacrificing manipulability. With our open-source hardware, we hope to see more robots that are soft and comfortable to interact with.
\section{Related Work}
Tactile sensing is becoming increasingly important in robotic systems, enabling robots to execute tasks such as object recognition \cite{bottcher2021object}, grasping \cite{guo2017robotic}, and uneven terrain locomotion \cite{wu2019tactile} using small-area tactile sensors. These sensors also facilitate whole-arm \cite{grice2013whole, jain2013reaching} and whole-body \cite{chengskincontrol, goncalves2022punyo} manipulation tasks enabled by large-area tactile sensors.
\subsection{Small-area tactile sensors}
Prior research has explored various sensing principles for building small-area tactile sensors. Among these are magnetic-based sensors, introduced by \cite{jamone2015highly, paulino2017low, donlon2018gelslim, yan2021soft, hellebrekers2019soft, bhirangi2021reskin}. These sensors detect force changes by tracking magnetic field fluctuations using magnetometers. They exhibit high precision, sensitivity, and resolution and can be made soft by incorporating magnetic particles into silicone gel. Vision-based sensors proposed by \cite{yuan2017gelsight, lambeta2020digit, kuppuswamy2020soft, zhang2021tactile} employ RGB cameras or other imaging devices behind a translucent membrane. This setup captures tactile feedback as the membrane deforms. Such sensors offer the advantages of high spatial resolution and resistance to electrical disturbances. Capacitive sensors developed in \cite{cotton2009multifunctional, Cap_Ji, LeMultimodalCap, XiongCap} function by measuring electric field changes, typically arising from variances in electrode distances. They are highly sensitive and incorporate soft materials for electrode molds. Similarly, fluid-based sensors \cite{biotac, fluid2013, kim2018soft, shi2016egain} detect force changes by sensing voltage variations arising from alterations in the cross-sectional area of the liquid channel under applied force. While small-area tactile sensors have demonstrated potential for extension to large areas, they face scalability issues, primarily due to intricate and costly fabrication processes \cite{yan2021soft, LeMultimodalCap}. Some of these sensors are cost-efficient and easy to fabricate, but their rigid and bulky design \cite{yuan2017gelsight, donlon2018gelslim, lambeta2020digit} renders them less suitable for pHRI applications that require whole-arm tactile sensing.
\subsection{Large-area tactile sensors}
Developing large-area tactile-sensing skins employs similar principles to small-area sensors and introduces methods apt for large-area sensing. Piezo-resistive skin sensors, as discussed in \cite{inaba1996full, 825396, mukai2008development, yoshikai2009development, wade2017force, luo2021learning, ye2022soft}, detect force changes via voltage variations from piezoelectric materials under pressure. Although these sensors adapt to curved robot surfaces, many either lack the desired softness or the passive compliance crucial for pHRI \cite{mukai2008development, ye2022soft, mittendorfer2011humanoid}. Capacitance-based tactile sensors, presented in \cite{ulmen2010robust,largeCapPatch,LargeCap2}, offer cost benefits, but existing sensors are not as stretchable. Air-inflated skins, proposed in \cite{goncalves2022punyo, alspach2015design,kim2019design}, monitor air pressure changes to detect forces. While they provide passive compliance, they have low spatial resolution and do not cover joints well. Vision-based large-area skins \cite{van2021large, luu2023simulation} tackle the issue of data management by mounting two cameras per link (top and bottom) to track marker displacements on a soft mold wrapped around the link. While these skins exhibit high spatial resolution and are soft, they fall short in joint coverage and lack the cushioning essential for comfortable pHRI. Current research has not yet fully addressed all the demands of pHRI, including softness, stretchability, comfort, and spatial resolution, in a comprehensive manner. Our proposed research seeks to address this gap.

\section{Skin Design and Fabrication}

In this section, we detail the design and fabrication for CushSense. We describe the construction of an individual sensing unit, known as a taxel (Sec. \ref{sec:single_taxel}). We also present the configuration of the skin (Sec. \ref{sec:whole_skin}), covering the entire robotic arm. This modular approach allows for easy adaptation, maintenance, and replacement. The design and circuit board files, fabrication details, material costs, and calibration utilities for this skin are open source \cite{cushsense23}.
\subsection{Building a Single Taxel}

A single taxel comprises nine layers (see Fig. \ref{fabrication process}). Table \ref{table: Taxel materials} lists the materials used to construct a taxel.

\label{sec:single_taxel}
\begin{table}[ht]
\renewcommand{\arraystretch}{1.25}
\begin{tabularx}{\columnwidth }{>{\raggedright\arraybackslash\hsize=.5\hsize\linewidth=\hsize}X >{\raggedright\arraybackslash\hsize=1\hsize\linewidth=\hsize}X >{\raggedright\arraybackslash\hsize=.9\hsize\linewidth=\hsize}X >{\raggedright\arraybackslash\hsize=1.6\hsize\linewidth=\hsize}X}
\hline
Layer & Material & Function & Composition \\ \hline
$\mathbf{A}$ & Stretch Conductive Fabric & Passive shielding and Grounding &  76\% Silver-coated Nylon, 24\% Elastic Fiber Fabric \\ \hline
$\mathbf{B}$ & Nylon Spandex Fabric & Spacing & 80\% Nylon, 20\% Spandex \\ \hline
$\mathbf{C}$ & Purple{\textregistered} Squishy & Dielectric & GelFlex elastic polymer, Polyurethane foam \\ \hline
$\mathbf{D}$ & Silverell Fabric & Conductive fabric & 16\% Silver-coated Nylon, 84\% Rayon \\ \hline
$\mathbf{E}$ & Silverell Fabric & Active shielding & 16\% Silver-coated Nylon, 84\% Rayon \\ \hline
\end{tabularx}
\caption{Taxel materials}
\label{table: Taxel materials}
\end{table}

 $\mathbf{A}$ layers act as a passive shield. $\mathbf{B}$ layers function as insulators to prevent shorting of the taxel. A thick, squishy dielectric layer, $\mathbf{C}$, sits sandwiched between two $\mathbf{D}$ layers of light, conductive fabric. These $\mathbf{D}$ layers serve as parallel plate electrodes in the capacitor. Layer $\mathbf{E}$ provides an active shield. Upon contact, taxel deformation causes a change in capacitance. Additionally, the design includes active and passive shielding to counteract parasitic capacitance and external interference, reducing sensor noise.

We fixed the taxel size to 3cm x 3cm for whole-arm skin. We conducted a series of characteristic experiments (Sec. \ref{sec. charac exps}) to choose the size, material, and design that optimizes the sensor performance for pHRI applications.

\subsection{Whole-arm Skin for a Robot Arm}
\label{sec:whole_skin}
We combined individual taxels with Capacitance to Digital Converters (CDCs) and custom-printed circuit boards (PCBs) to create a tactile-sensing skin for the Kinova Gen3 7-DoF robot arm. We employed a modular design approach, ensuring users can easily configure taxels to cover specific sections of interest. If taxels malfunction, users can replace them by removing the old ones and sewing new ones on $\mathbf{B}$.

We created four sections of our proposed sensor, CushSense, for the links that are most likely to contact users when providing assistance \cite{grice2013whole}. Each section comprises anywhere from 9 to 19 taxels, depending on the link size. Fabricating the taxels involved cutting the fabric and sewing the layers together. We crimped the signal wires with metal loop connectors and sewed them to the conductive fabric layers with conductive thread. We used E6000 fabric glue to stick the layers together. To create these sections, we laser-cut $\mathbf{A}$ and $\mathbf{B}$ and sewed them together while connecting $\mathbf{A}$ to the ground as passive shielding. We glued $\mathbf{C}$, $\mathbf{D}$, and $\mathbf{E}$ together and then sewed $\mathbf{E}$ onto $\mathbf{B}$ to stabilize the taxel in-place.

For each link, we integrated an I2C multiplexer (Adafruit TCA9548a) to simplify the data collection process from all taxels and to accommodate the skin's modularity. Each multiplexer can support up to 7 CDCs with the same address, and each CDC can support up to four taxels. We choose an off-the-shelf FDC1004 CDC, a 4-channel CDC board that offers a wide capacitance range of $\pm$ 15pF and a measurement resolution of 0.5fF. It provides output rates ranging from 100-400Hz. Furthermore, it integrates an active shield driver that reduces electromagnetic interference from the surrounding environment, enhancing the reliability of CushSense. Using a single Arduino Nano, our design can accommodate up to 224 taxels. Additional microcontrollers may be integrated depending on the application.



\section{Sensor Characterization of a Single Taxel}

\label{sec. charac exps}

\begin{figure}[!tbp]
    \vspace{5pt}
    \centering
    \begin{subfigure}{0.21\textwidth}
        \includegraphics[width=\linewidth]{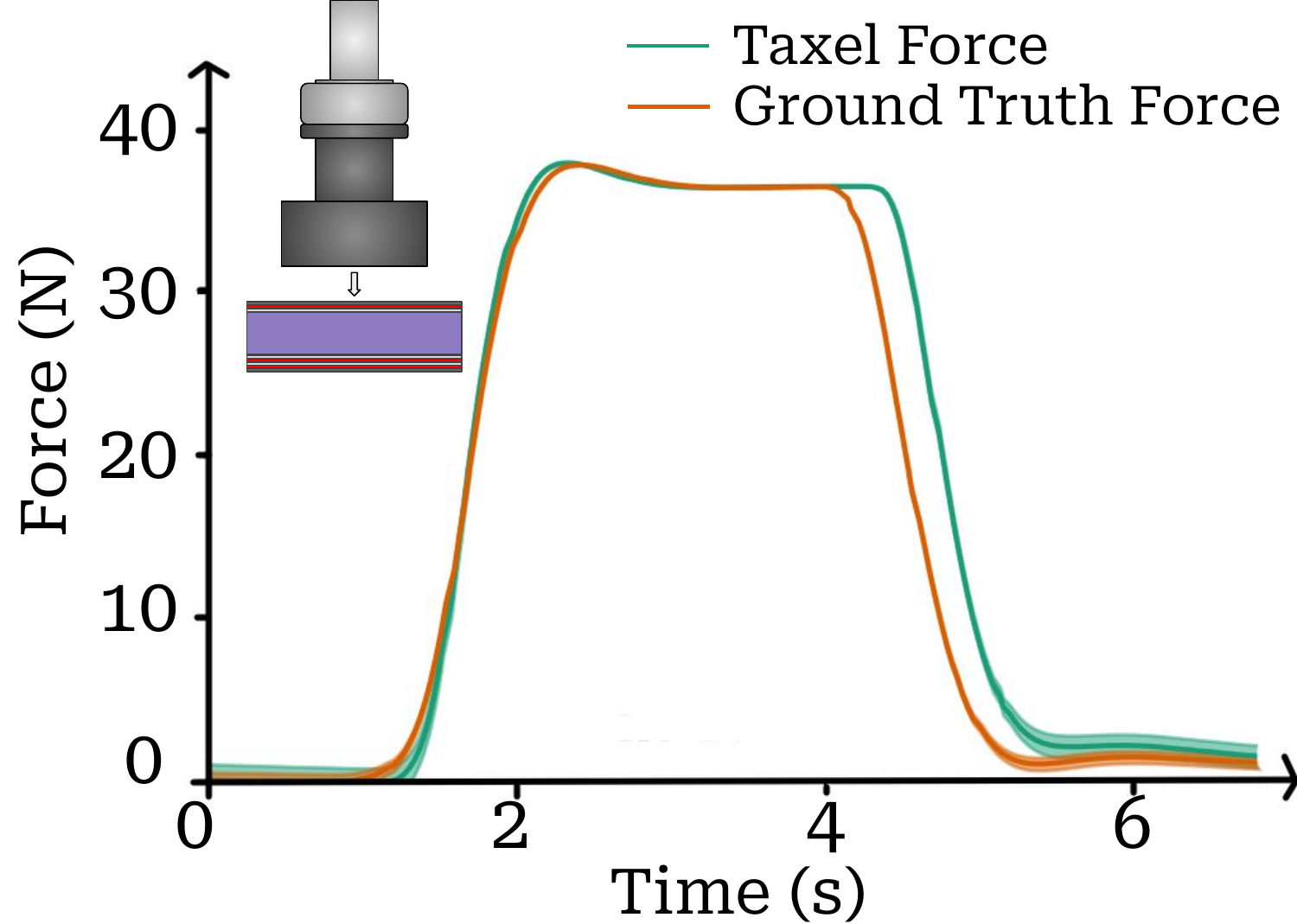}
        \vspace{-17pt}
        \caption{Accuracy}
        \label{Accuracy}
    \end{subfigure}
    \vspace{20pt}
    \begin{subfigure}{0.21\textwidth}
        \includegraphics[width=\linewidth]{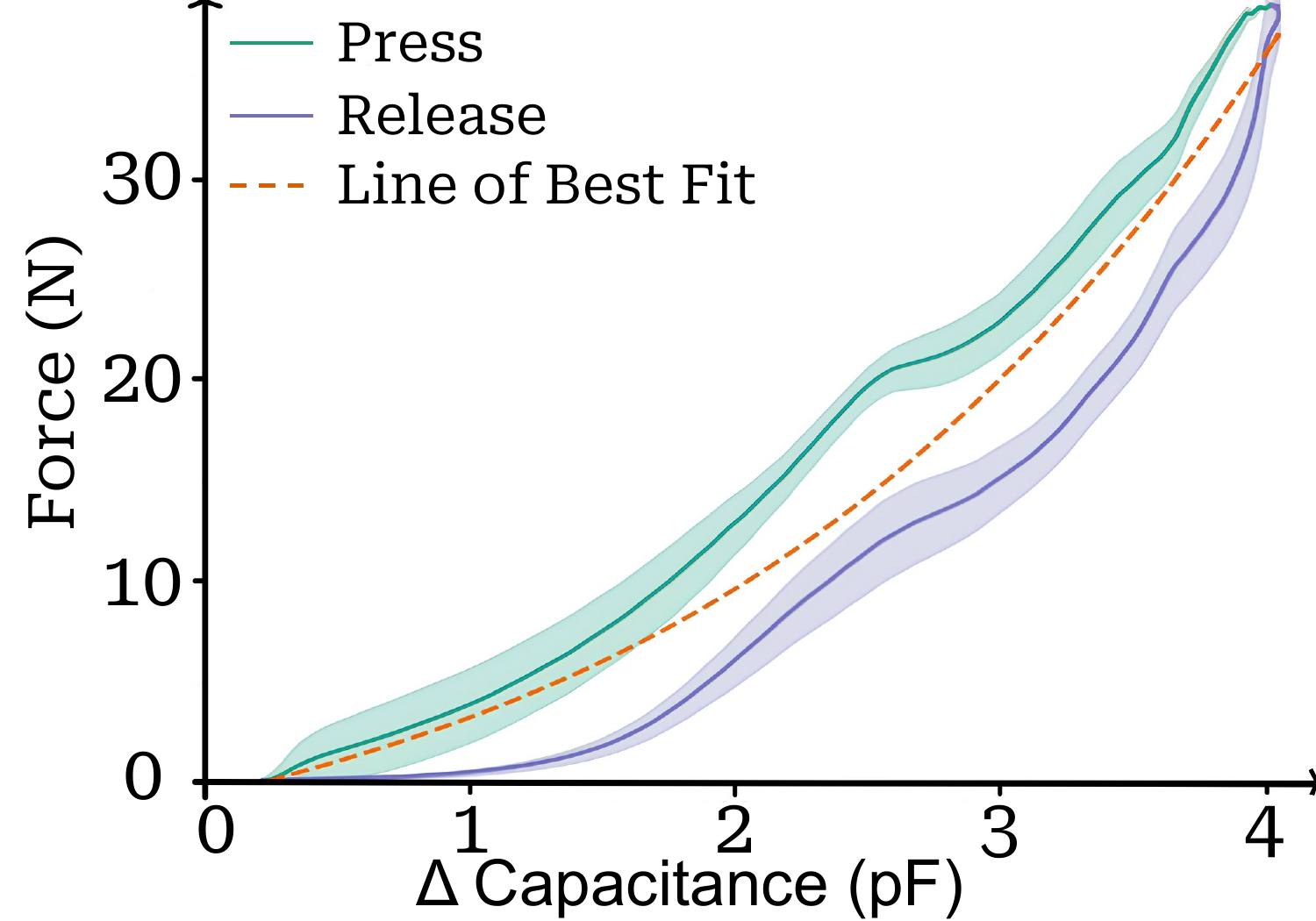}
        \vspace{-17pt}
        \caption{Hysteresis}
        \label{Hysteresis}
    \end{subfigure}
    \newline
    \begin{subfigure}{0.21\textwidth}
        \includegraphics[width=\linewidth]{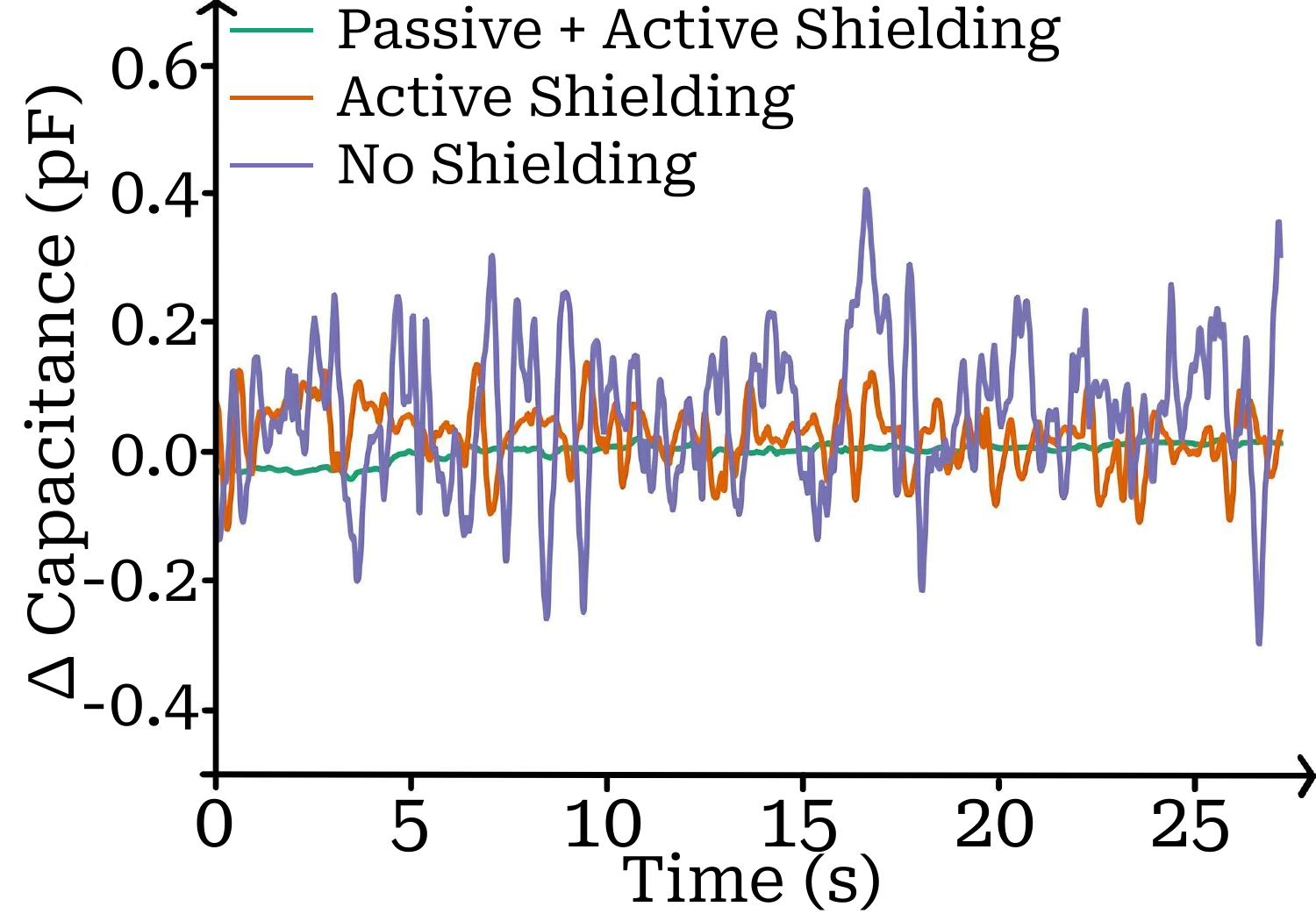}
        \vspace{-17pt}
        \caption{Noise Reduction}
        \label{Noise Reduction}
    \end{subfigure}
    \vspace{20pt}
    \begin{subfigure}{0.21\textwidth}
        \includegraphics[width=\linewidth]{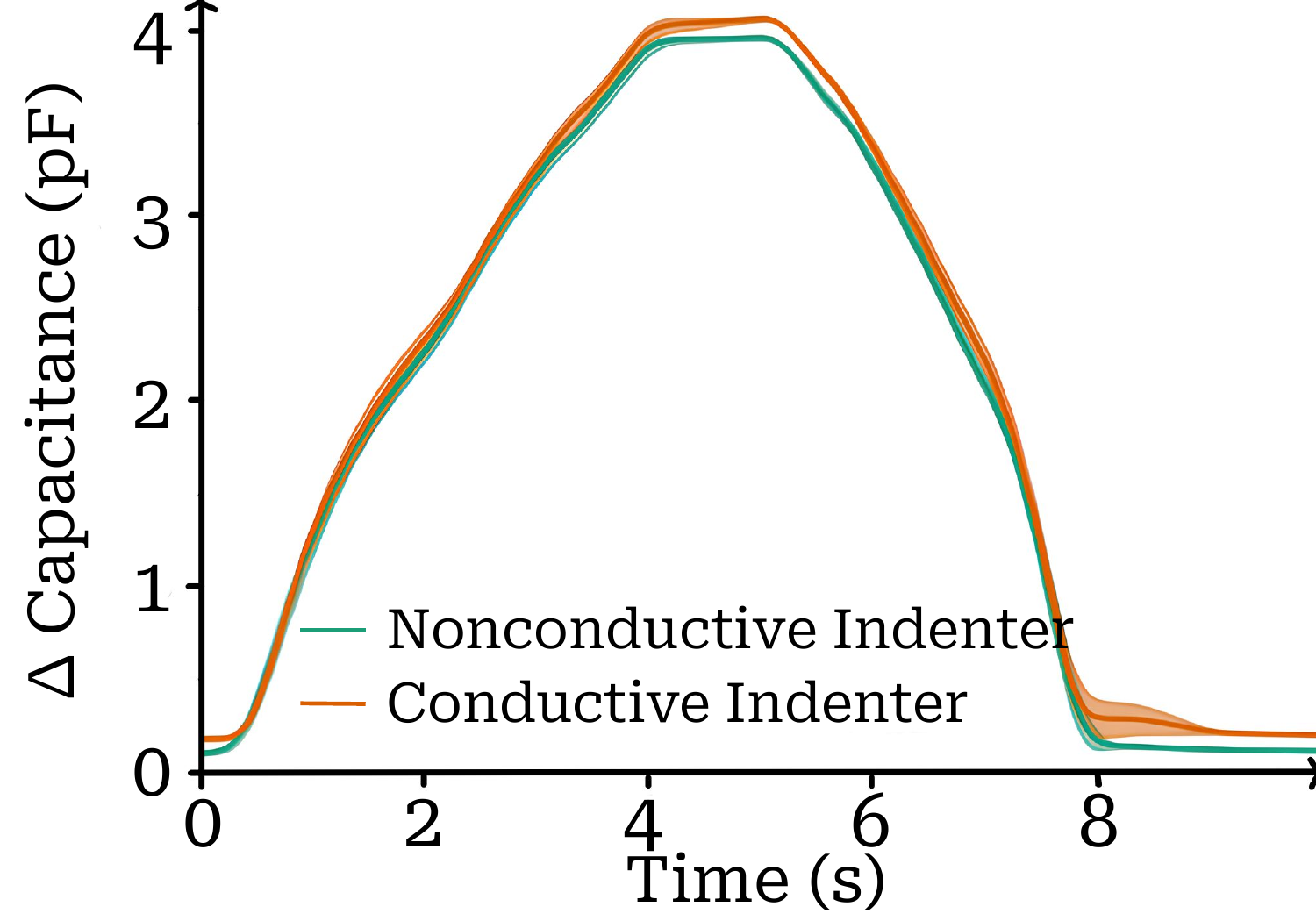} 
        \vspace{-17pt}
        \caption{Contact Testing}
        \label{Contact Testing}
    \end{subfigure}
    \begin{subfigure}{0.21\textwidth}
        \includegraphics[width=\linewidth]{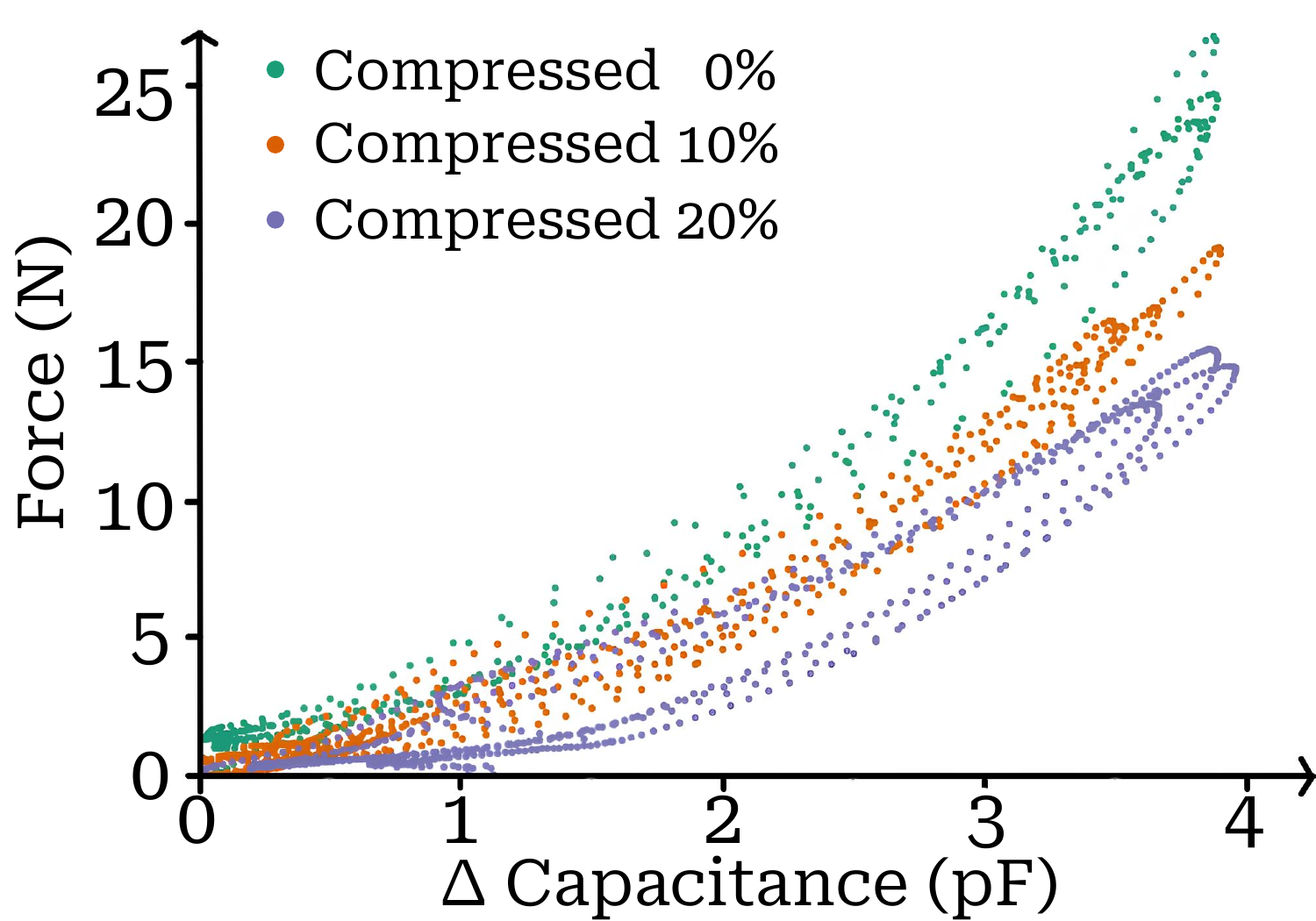} 
        \vspace{-17pt}
        \caption{Effect of Compression}
        \label{Effect of Compression}
    \end{subfigure}
    \vspace{10pt}
    \begin{subfigure}{0.21\textwidth}
        \includegraphics[width=\linewidth]{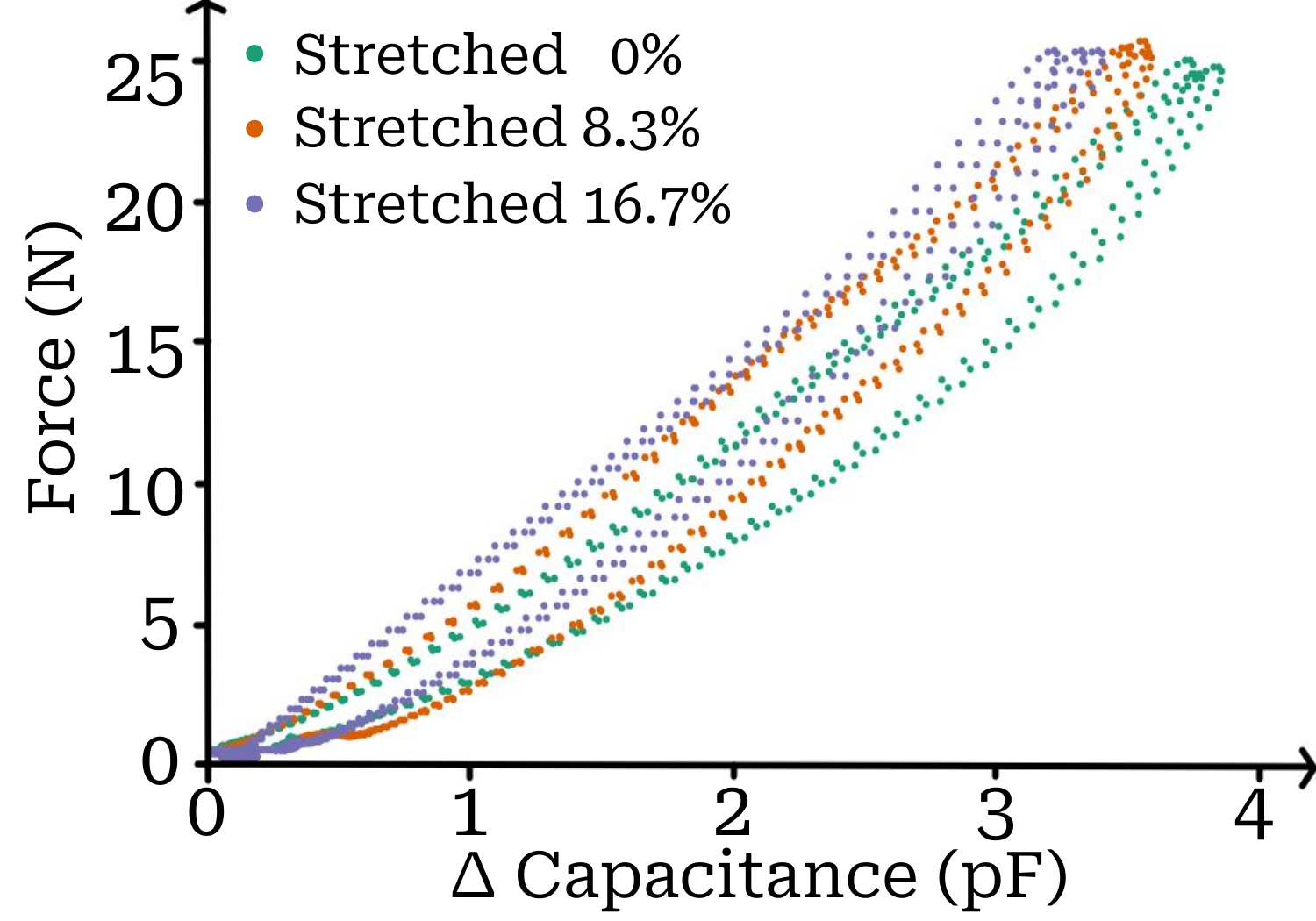} 
        \vspace{-17pt}
        \caption{Effect of Stretching}
        \label{Effect of Stretching}
    \end{subfigure}
    \caption{\textbf{Sensor Characteristics and Performance Evaluation.} The top-left corner of (a) shows the experiment setup comprising F\slash T Sensor (in light gray), Indenter (in dark gray), and the taxel (in purple). Experiment (a) validates the force-sensing capability of our skin, (b) exhibits minimal hysteresis error, (c)-(d) shows reduction of noise due to electromagnetic interference, and (e)-(f) retains sensitivity under compression and deformation.}
     \vspace{-8pt}
\end{figure}

\vspace{1pt}
This section provides a detailed evaluation of our soft and stretchable tactile sensor. We include an in-depth examination of its intrinsic characteristics and a comparative analysis of alternative designs.

\subsection{Sensor Characteristics and Performance Evaluation}
We used the ATI Nano25 Force Torque (F\slash T) sensor to measure the ground truth force during our experiments. Fig.\ref{Accuracy} shows the experiment setup. The average measurement range of our sensor is approximately 0-5 pF, which allows it to sense forces up to 55N.

\textbf{Sensing Accuracy}. To evaluate the accuracy, we compared the estimated force values with the ground truth measurements recorded by the F\slash T sensor. We applied forces ranging from 0N to 40N and recorded both force values over 10 press-release cycles. Results (see Fig. \ref{Accuracy}) show that our sensor exhibits a low relative error of $0.58\%$ with respect to the measuring range, making it ideally suited for pHRI applications requiring accurate force feedback.

\textbf{Hysteresis}. Hysteresis is the phenomenon in which the sensor output depends on the current input and the previous inputs. We conducted a hysteresis experiment to evaluate sensor reliability by performing 10 pressing and releasing force cycles. The hysteresis error (see Fig. \ref{Hysteresis}) of our sensor is 3N, about 5.4$\%$ of its measuring range. This error level is acceptable for whole-arm sensing for pHRI with ample hardware compliance like ours.

\textbf{Noise Reduction and Contact Testing}. Capacitive sensors are sensitive to changes in the electromagnetic field, especially near electronics and circuitry. To mitigate this interference, we employed both passive and active shielding. Our passive shielding uses highly conductive fabric to ground and isolate the sensor. Active shielding maintains potential equilibrium between the shield and the sensing layer by using the same voltage as the sensor input signal.

Fig. \ref{Noise Reduction} illustrates the noise reduction of the sensor. Without shielding, the sensor detected noise is 14.2$\%$ of the measuring range. With active shielding, the noise decreased to 5.2$\%$. The combination of active and passive shielding further reduced the noise to 3.2$\%$, which mitigates $77.5\%$ of the noise detected by the sensor without shielding. In addition to noise reduction, shielding enables our sensor to provide consistent readings for contacts of the same magnitude, regardless of whether they are conductive or non-conductive (see Fig. \ref{Contact Testing}).

\textbf{Effect of Compression}. We experimented with lateral compression to assess its impact on sensor performance. We performed the tests in three groups: no compression, 10\% compression, and 20\% compression. We measured the F\slash T sensor response and the sensor capacitance data in each group. As shown in Fig. \ref{Effect of Compression}, greater compression increases the sensitivity. However, our sensor retains adequate sensitivity for force sensing in configurations where the taxels around the joints experience lateral compression.

\textbf{Effect of Stretching}. We performed the tests for this experiment in three groups: no stretch, stretched 8.3$\%$, and 16.7$\%$. Lateral stretching can slightly reduce the sensor's sensitivity (see Fig. \ref{Effect of Stretching}). It retains the necessary sensitivity ideal for stable sensing around the joints of the robot.

\textbf{Durability Testing}. Tactile sensors should demonstrate reliable sensing upon extended use. We performed a durability test to validate this by probing a single taxel 1000 times. In the end, we observed a tiny drop of 0.054\% in force estimation accuracy compared to the initial accuracy of 98.80\%, demonstrating the sensor's robustness and ability to maintain accuracy over prolonged operational periods.

\subsection{Comparative Analysis of Alternative Designs}
\begin{figure}[t]
    \vspace{5pt}
   \centering
   \begin{subfigure}{0.9\columnwidth}
        \includegraphics[width=\linewidth]{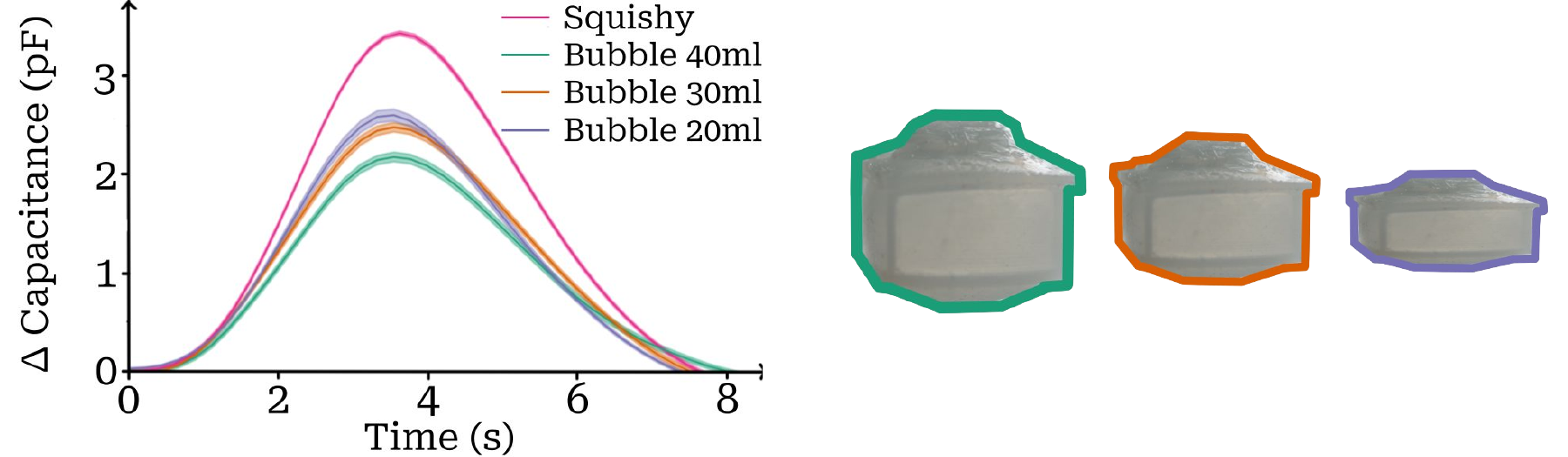}
        \vspace{-17pt}
        \caption{Bubble as Dielectric Material}
        \label{bubble}
    \end{subfigure}
    \vspace{15pt}
    \newline
    \centering
    \hspace{-0.1cm}
    \begin{subfigure}{0.9\columnwidth}
        \includegraphics[width=\linewidth]{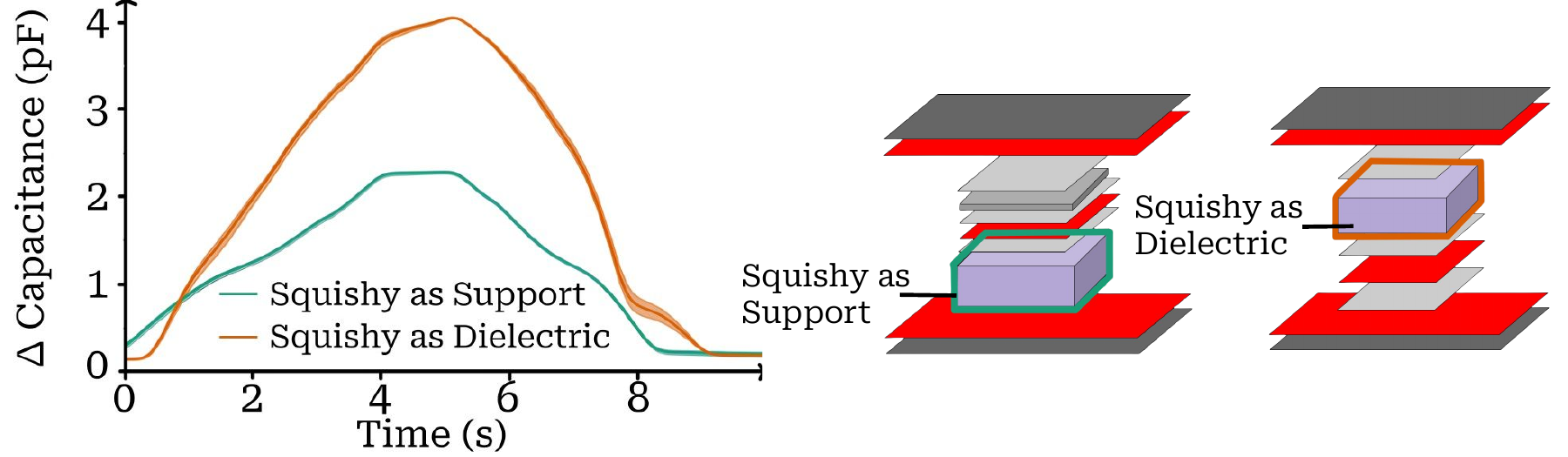}
        \vspace{-17pt}
        \caption{Role of Squishy Material}
        \label{support}
    \end{subfigure}
    \vspace{15pt}
    \newline
    \vspace{8pt}
    \centering
    \hspace{-0.5cm}
    \hspace{-0.3cm}
    \begin{subfigure}{0.9\columnwidth}
        \includegraphics[width=\linewidth]{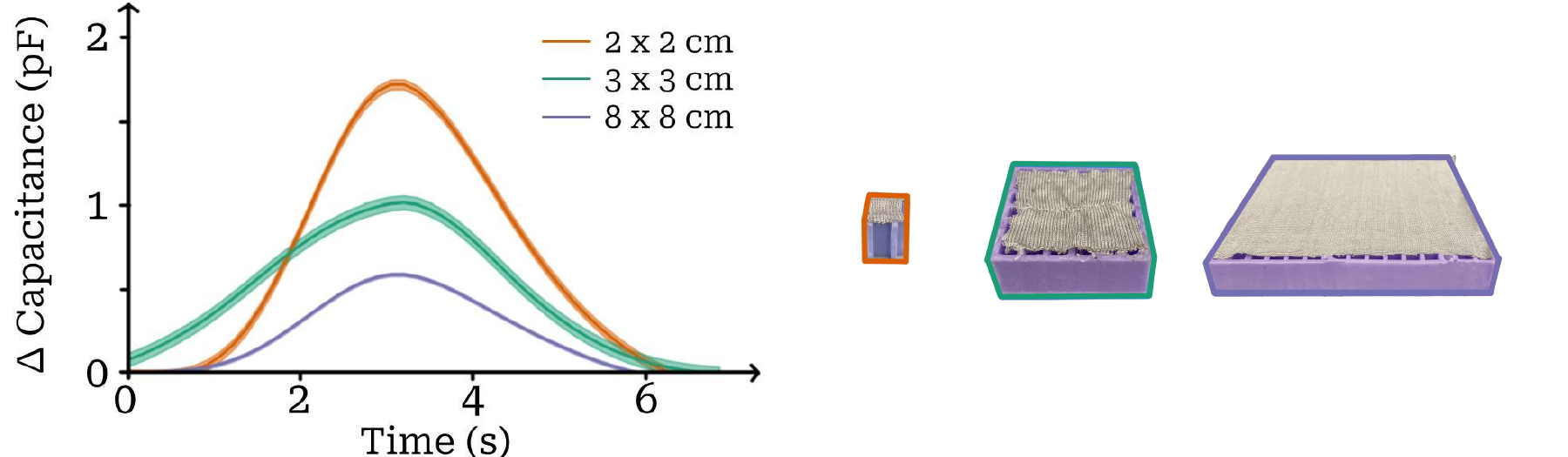}
        \vspace{-17pt}
        \caption{Different Size of A Taxel}
        \label{size}
    \end{subfigure}
    \caption{\textbf{Comparative Analysis with Alternative Designs}. Experiments (a)-(b) illustrate that Squishy as a dielectric maximizes sensitivity. (c) Smaller taxels exhibit further increases in sensitivity. We chose a taxel size of 3x3 cm based on a tradeoff between sensitivity and ease of construction.}
    \vspace{-12pt}
\end{figure}
We also conduct a comparative analysis of our sensor design against alternative designs to evaluate its performance in terms of accuracy, stability, and sensitivity.
\begin{figure*}[ht!]
    \vspace{5pt}
    \centering
    \begin{subfigure}{0.23\textwidth}
        \centering
        \includegraphics[width=\textwidth]{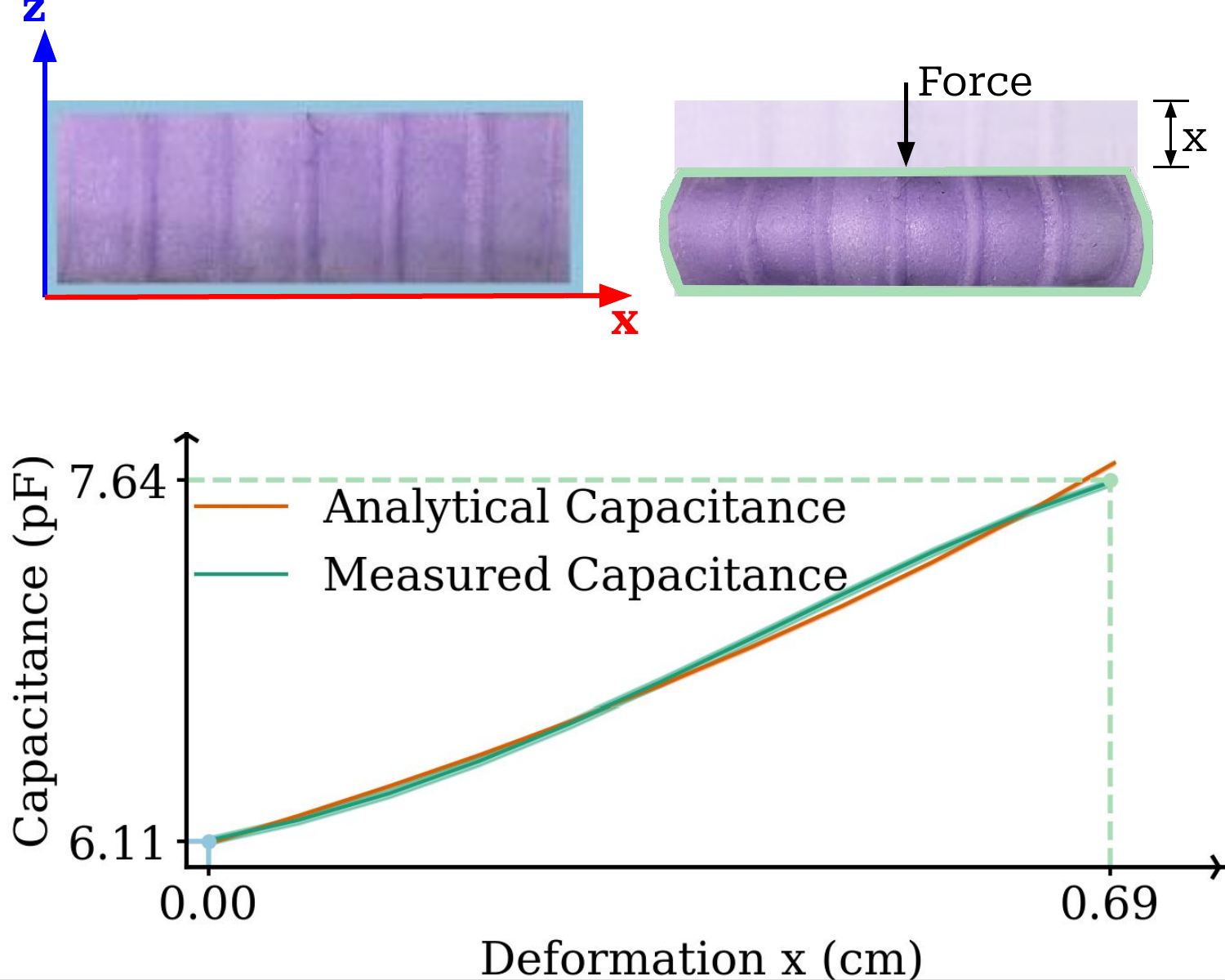}
        \caption{\strut Axial Compression}
        \label{Axial Compression}
    \end{subfigure}
    \begin{subfigure}{0.23\textwidth}
        \centering
        \includegraphics[width=\textwidth]{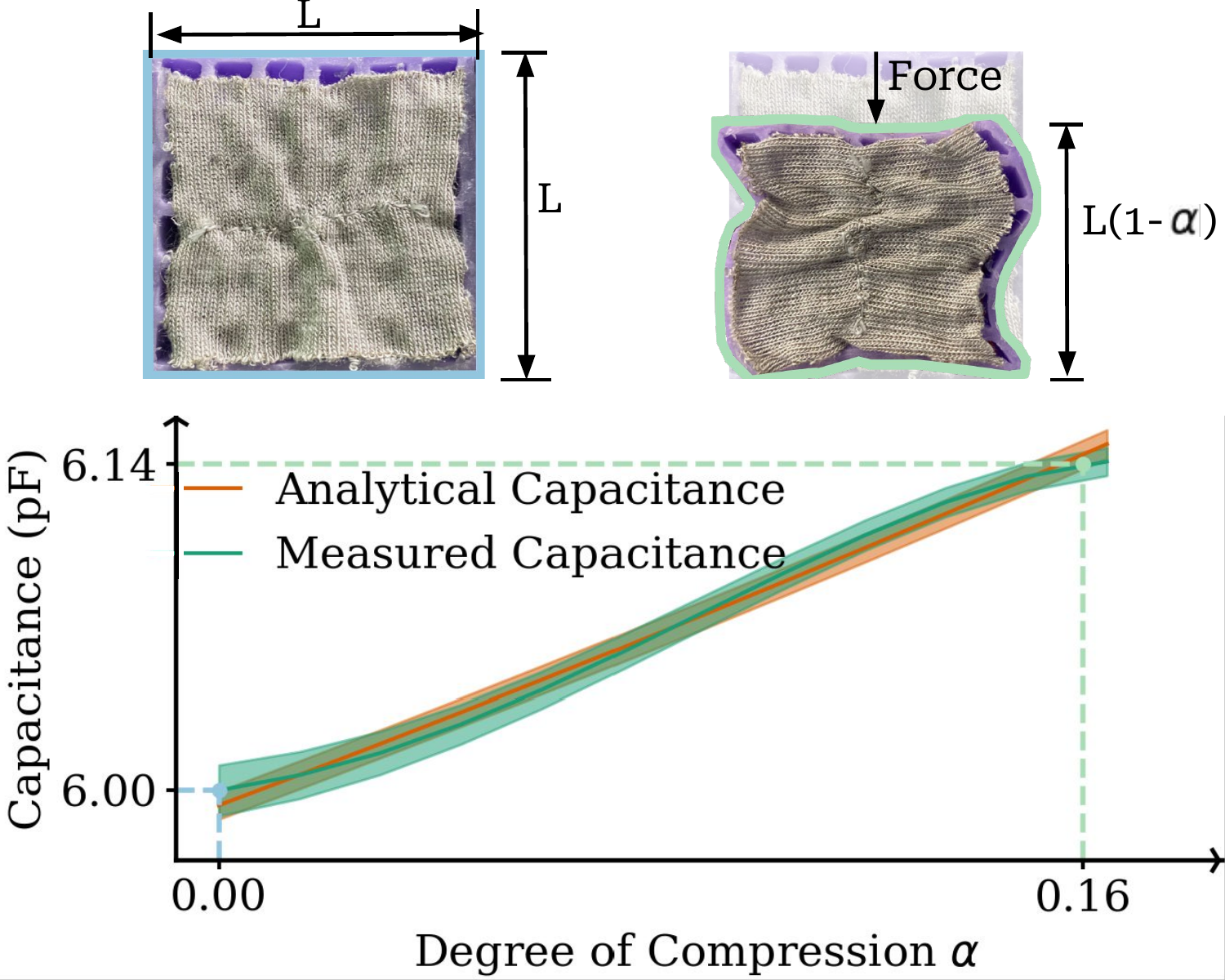}
        \caption{\strut Lateral Compression}
        \label{Lateral Compression}
    \end{subfigure}
    \begin{subfigure}{0.23\textwidth}
        \centering
        \includegraphics[width=\textwidth]{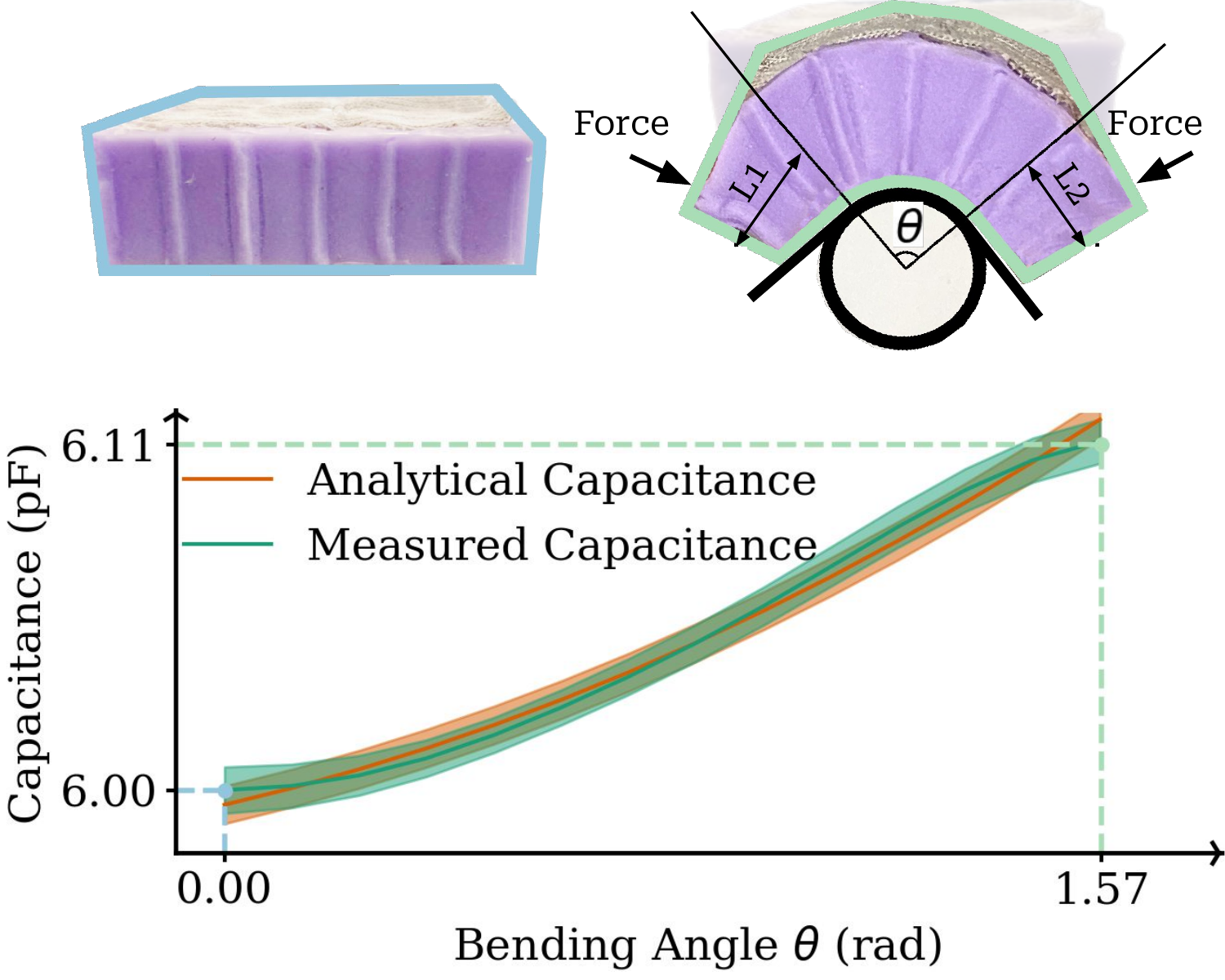}
        \caption{\strut Bending}
        \label{Bending}
    \end{subfigure}
    \begin{subfigure}{0.23\textwidth}
        \centering
        \includegraphics[width=\textwidth]{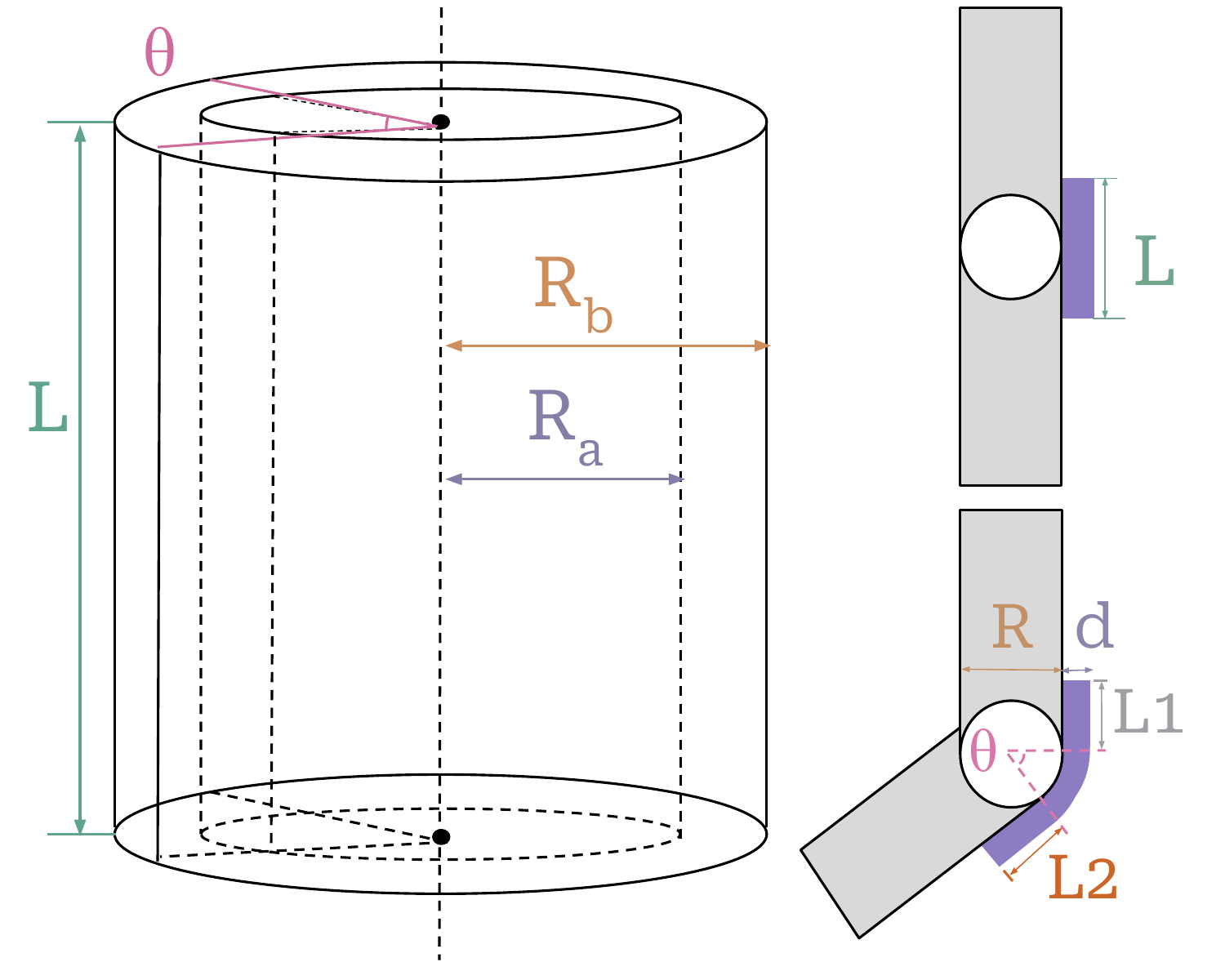}
        \caption{\strut Taxel Bending Model}
        \label{bendmodel}
    \end{subfigure}
    \vspace{10pt}
    \caption{\textbf{Math Model Validation.} Comparison plots validate the efficacy of the proposed mathematical model for different actions, where \textit{Measured Capacitance} represents experimental data, and \textit{Analytical Capacitance} is based on the developed mathematical model. The fitted model closely resembles the measured data for all actions.}
    \label{Experiments for Math Model}
\vspace{-4pt}
\end{figure*}

\textbf{Material of Dielectric Layer}. We compared a soft bubble filled with air and Squishy (see Table \ref{table: Taxel materials} as the dielectric layer. As shown in Fig. \ref{bubble}, Squishy exhibited the highest change in capacitance $\Delta$C, indicating better sensitivity than the sensor with a bubble. 
Also, occasional air leaks reduce its cushioning and sensitivity over time when using the bubble as the dielectric layer. As shown in Fig. \ref{bubble}, we tested the sensor response when inflating the bubble with 40ml, 30ml, and 20ml of air. As the bubble deflates, the capacitance value increases, leading to a discrepancy in the calibration function.

\textbf{Role of Squishy}. We evaluated sensor performance with Squishy (\textbf{C}) acting both as the dielectric layer and as a soft support. Each configuration provides a cushioning effect. However, for identical deformations, when Squishy is the dielectric layer (refer to Fig. \ref{support}), the change in capacitance ($\Delta$C) is nearly twice that compared to its role as a mere support. This increase is due to the Scuba Fabric \cite{scuba} used for support being much thinner than Squishy (6 times less thick), resulting in a smaller distance change between electrodes and, thus, a smaller $\Delta$C for an applied force. Thus, a thicker soft material as the dielectric layer notably enhances sensor sensitivity by yielding a larger variation in sensor readings.

\textbf{Size of a taxel}. The size of a taxel directly influences the spatial resolution of our tactile-sensing skin. Opting for smaller taxels improves resolution but introduces more noise and complicates the fabrication process (see Fig. \ref{size}). The sensitivity of 2cm $\times$ 2cm taxel size is almost three times that of 8cm $\times$ 8cm taxel size. However, its noise range is about 0.04pF, about four times the 8cm $\times$ 8cm taxel. Also, smaller taxels increase fabrication complexity and require more circuitry for the same area coverage. So, after weighing sensitivity, resolution, noise reduction ability, and fabrication complexity, we finally chose 3cm $\times$ 3cm as the size of a taxel for our skin.

\section{Capacitance Modeling And Validation}

We formulated a mathematical model to predict how various actions, including Axial Compression, Lateral Compression, and Bending, impact the capacitance of our taxel-based sensor. Validation of this model involved conducting a series of experiments, as illustrated in Fig. \ref{Experiments for Math Model}. This model aids in confirming the proper fabrication of taxels using the analytical values in the absence of an expensive sensor for ground truth data. Additionally, modeling the effects of Lateral Compression and Bending allows us to offset their corresponding capacitance changes for accurate sensing around the joints when the robot is in motion. Each experiment comprised 10 trials. The basis of our mathematical model is the capacitance formula, $C = \dfrac{\epsilon A}{d}$, \vspace{2pt}where $C$ represents capacitance, $\epsilon$ relative permittivity, $A$ overlapped electrode area, and $d$ the distance between electrodes.

\textbf{Axial Compression}. In this scenario, force is applied to the top surface of the taxel. Since the cross-sectional area undergoes negligible change during axial compression, we assume that the area $A = L*L$ and the permittivity $\epsilon$ remain constant. The compression causes the thickness of the dielectric layer $d$ to decrease. Therefore, the new capacitance $C_1$ can be expressed as $C_0\dfrac{h_0}{h_1}$. \vspace{2pt} We use subscript `$0$' to denote the initial state and subscript `$1$' to denote the final state.
From the equation, it is clear that the capacitance and thickness of the taxel after compression are inversely proportional. We present the experimental result in Fig. \ref{Axial Compression}. The fitted mathematical model is:
\begin{equation}
C_1 = 5.88\dfrac{1}{h_1} + 2.16 = 5.88\dfrac{1}{h_0-x} + 2.16
\end{equation}

where x is the deformation. We obtained this analytical form using Ordinary Least Squares (OLS) linear regression. We assumed a fixed standard deviation for the analytical values, modeling it as 0.1\% of the analytical capacitance.

\textbf{Lateral Compression}. Assuming horizontal forces act on the taxel and cause it to compress by a ratio of $\alpha$, we express the new taxel area as $A_1 = L * L * (1 - \alpha) = A_0(1 - \alpha)$. Due to the grid-like structure of the dielectric layer, compressing the taxel reduces the proportion of air and increases the proportion of rubber, resulting in an overall increase in the permittivity of the dielectric layer. We hypothesize that this change in permittivity is linearly related to the compression ratio: $\epsilon_1 = (1 + m\alpha)\epsilon_0$, where $m$ is a positive proportionality constant. Lateral compression of the taxel by 50\% causes the distance between the electrodes to increase by just 1 mm, approximately 6\% of the original distance. So, we assume that $h_1 = h_0$. Consequently, we establish:
\begin{equation}
    \begin{split}
         C_1 &= \dfrac{\epsilon_0(1 + m\alpha)A_0(1 - \alpha)}{h_0} \\
         &= \dfrac{\epsilon_0A_0}{h_0}(1 + m\alpha - \alpha - m\alpha^2) 
    \end{split}
\end{equation}
We can observe that $C_1$ is a quadratic function of $\alpha$. The experimental result is shown in Fig. \ref{Lateral Compression}. The fitted mathematical model is: $C_1 = 0.44\alpha^2 + 0.87\alpha + 5.99$, obtained using second-order polynomial regression.

\textbf{Bending}. When positioned between two links, the taxel bends as the joint connecting them rotates. Our model, shown in Fig. \ref{bendmodel}, divides the capacitance into bent and linear regions.

\begin{figure}[ht!]
    \centering
    \begin{subfigure}{0.97\columnwidth}
        \centering
        \includegraphics[width=\textwidth]{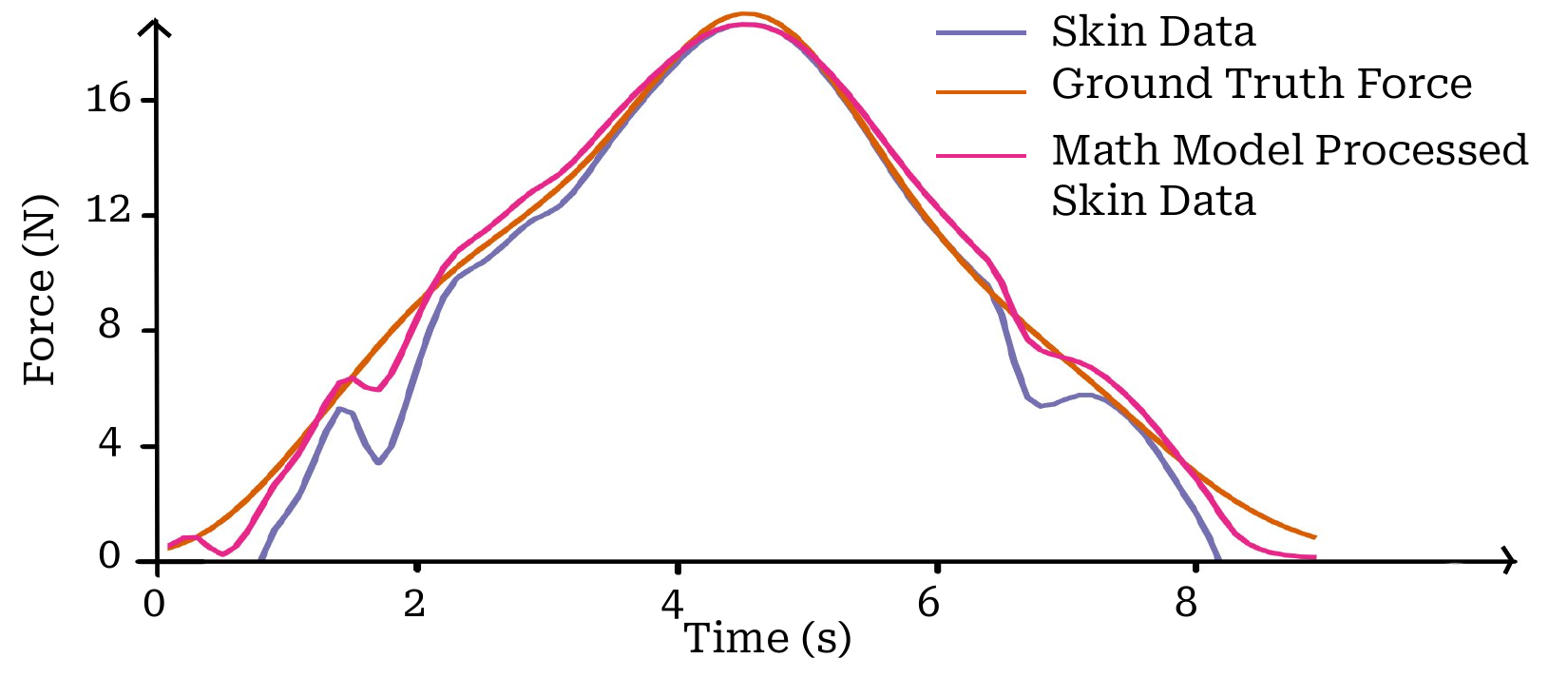}
    \end{subfigure}
    \vspace{-5pt}
    \caption{\textbf{Model-based Filtering.} Comparison of ground truth forces with skin data and the math model processed data. The processed data is obtained using \textit{filtered} = 0.4 * \textit{analytical} + 0.6 * \textit{measured}, where \textit{analytical} is calculated using the math model and \textit{measured} is the skin data. The processed data demonstrates reduced noise and aligns well with ground truth.}
    \label{refmathvalue}
    \vspace{-11pt}
\end{figure}

\begin{figure*}[t!]
    \vspace{5pt}
  \centering
  \begin{subfigure}{0.93\textwidth}
    \centering
      {\includegraphics[width=\textwidth]{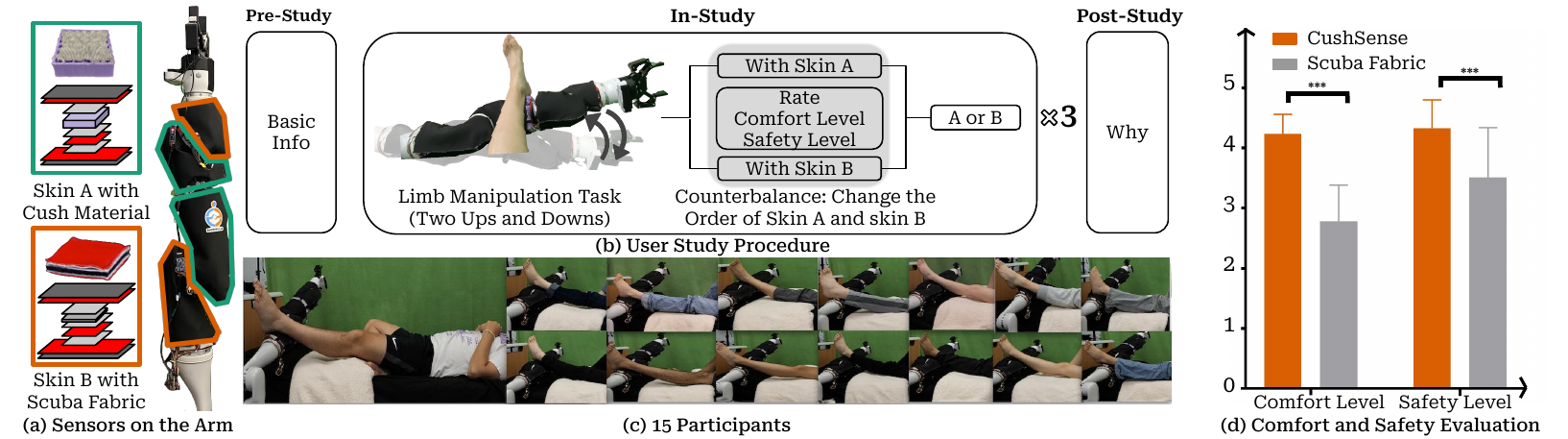}}
  \end{subfigure}
  \caption{ \textbf{User Study and Analysis.} Experimental setup, study procedure, and results from the limb manipulation user study evaluating the two skins' perceived safety and comfort. Most participants preferred the Cush Material over the Scuba Fabric. *** indicates significant differences with $p_{0.0001}$.}
  \label{fig:user_study}
  \vspace{-7pt}
\end{figure*}

In the linear region, the taxel is firmly fixed on the joint so that the positions of its two ends on the link remain unchanged, and thus $A_{L} = A_{L1} + A_{L2} = A_0$. When the taxel bends, the dielectric layer stretches, leading to an increase in the proportion of air within the dielectric layer and reducing its overall permittivity. We hypothesize that this reduction is linearly related to the angle of bending $\theta$, resulting in $\epsilon_1 = (1 - m\theta)\epsilon_0$. Additionally, due to the grid-like structure and the stretchable characteristic of the dielectric layer, the distance between the electrodes remains almost unaffected when we bend the taxel. Therefore, we assume $h_1 = h_0$. Consequently, the capacitance of the cumulative linear region can be expressed as $C_L = \dfrac{\epsilon_0(1 - m\theta)A_0}{h_0} = C_0(1 - m\theta)$.
When considering the bent region, we use a cylindrical capacitor model, as shown in Fig. \ref{bendmodel}. This model represents a cylindrical capacitor of length $L$ formed by an inner and an outer conductive cylindrical shell with radii $R_a$ and $R_b$, respectively. The capacitance of the cylindrical capacitor can be expressed as $C = \dfrac{2\pi\epsilon L}{ln(R_b/R_a)}$.\vspace{2pt}
In case of the capacitance of the bent region, $R_b = R_a + d$, $\epsilon = (1 - \theta)\epsilon_0$, and the capacitance only takes $\theta/2\pi$ of the capacitor value of the entire cylindrical model. So, the capacitance of the bending region is:
\begin{equation}
C_B = \dfrac{2\pi\epsilon_0(1 - m\theta) L}{ln(1+d/R_{a})}\dfrac{\theta}{2\pi} = \dfrac{\epsilon_0 L}{ln(1 + d/R_{a})}(\theta - m\theta^2)
\end{equation}

Thus, the total capacitance of the taxel would be the sum of $C_B$ and $C_L$:
\begin{equation}
\begin{split}
C_{total} & = C_B + C_L \\
        & = C_0(1 - m\theta) + \dfrac{\epsilon_0 L}{ln(1 + d/R_{a})}(\theta - m\theta^2) \\
        & = a\theta^2 + b\theta +c
\end{split}
\end{equation}

where $a, b, c$ are proportionality constants given by the physical parameters of the taxel. The experimental result is shown in Fig. \ref{Bending}. The fitted mathematical model is: $C_1 = 0.024\theta^2 + 0.041\theta + 5.99$.=, obtained using second-order polynomial regression.

\section{User Study}

Providing physical assistance for ADLs, such as dressing, bathing, and transferring, often involves limb manipulation \cite{repositioning, madan2022sparcs} as a critical task. To evaluate the perceived safety and comfort of our tactile skin during pHRI, we centered our user study around limb manipulation. We recruited 15 human subjects (9 Female, 6 Male) to participate in our study. Their mean age was 22.27 years (SD = 4.11), and their mean weight was 137.73 lb (SD = 32.98). Of all the participants, 40\% had previous experience with robots.

\subsection{Experimental Setup}
The main source of comfort in CushSense is the dielectric material. To demonstrate how it affects perceived safety and comfort, we compared our sensor with an alternate version, where we replaced the dielectric layer with Scuba Fabric (see Fig. \ref{fig:user_study}). We chose Scuba Fabric since it is soft and stretchable, yet does not offer the same degree of cushioning. During the study, the robot arm was positioned under the ankle of the participants.

\subsection{Experimental Procedure}
We simulate the task of limb manipulation as it would occur in assisting with dressing a user's lower body. We ask participants to lie on a bed with their eyes blindfolded (to eliminate visual bias). We place their ankles against one of the two skins. The robot arm then raised their left leg, maintained the position for 5 seconds, and gently lowered it. This process was repeated twice for each skin. The participants then evaluated the comfort and safety of each skin and rated them based on the following criteria:
\begin{itemize}
\item \textbf{Comfort level}: Participants rated their perceived comfort on a five-point scale ranging from \textit{very comfortable} to \textit{very uncomfortable}.
\item \textbf{Safety level}: We assessed the perceived safety using a similar five-point scale ranging from \textit{very safe} to \textit{very unsafe}.
\end{itemize}

The above procedure was repeated three times with the skin variants (CushSense and Scuba Fabric) counterbalanced across users. Detailed information on the user study questionnaire can be found on our website \cite{cushsense23}. The study protocol was reviewed and approved by the Institutional Review
Board at Cornell University (Protocol \#IRB0146211). All participants provided written informed consent to participate in the study.

\begin{figure}[t!]
    \centering
    \begin{subfigure}{0.79\columnwidth}
        \centering
        \includegraphics[width=\textwidth]{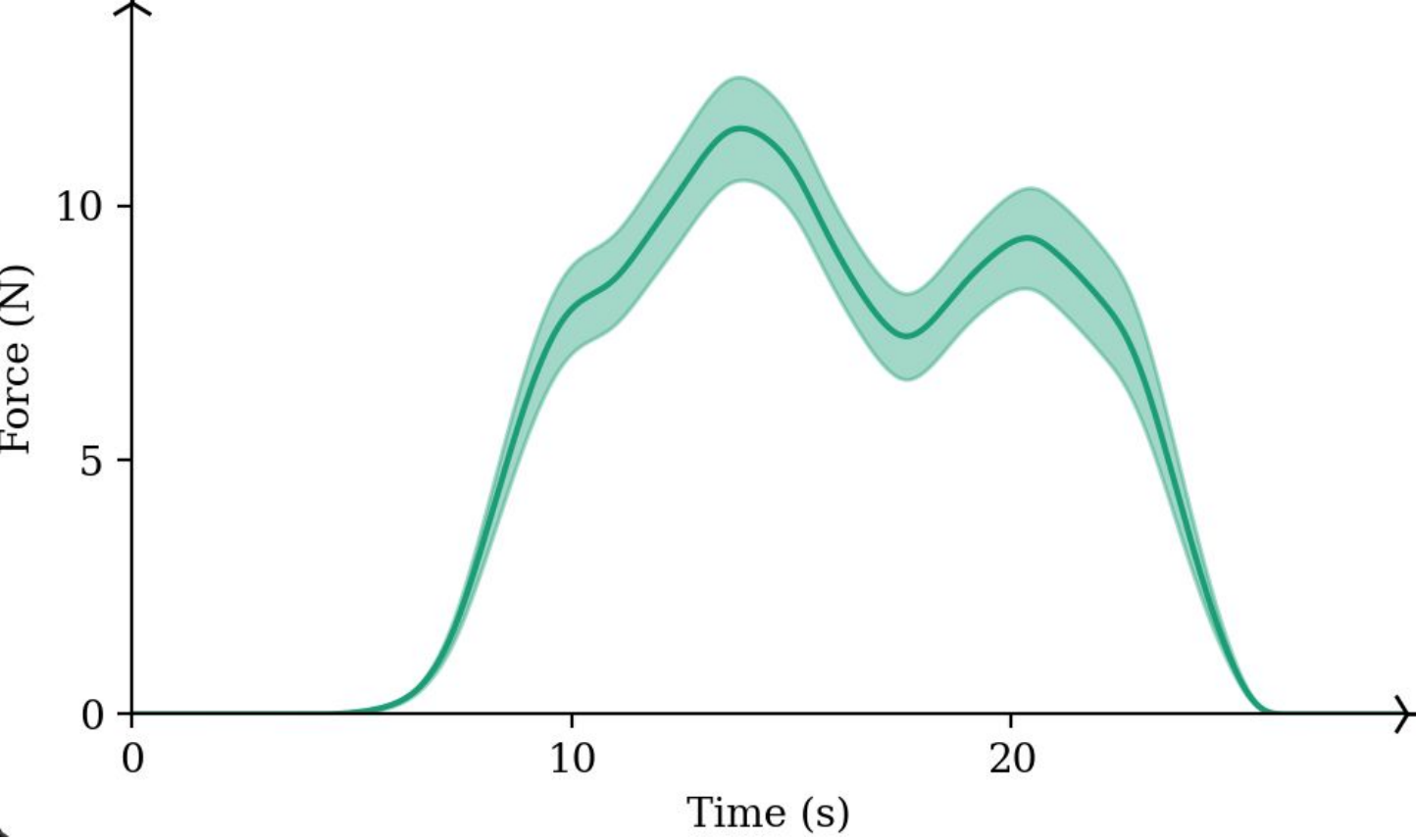}
    \end{subfigure}
    \vspace{-5pt}
    \caption{CushSense forces averaged across three trials for all participants.}
    \label{forceplot}
    \vspace{-11pt}
\end{figure}

\subsection{Results and Discussion}
14 out of 15 participants preferred CushSense over the Scuba Fabric version. 
A summarized comparison is presented in Fig. \ref{fig:user_study}, with force values shown in Fig. \ref{forceplot}. Overall, the evaluations highlight a consistent inclination towards the comfort and safety offered by CushSense compared to the Scuba Fabric. A paired sample t-test conducted on the comfort ($t_{14}$=7.995, p$<$0.001) and safety ($t_{14}$=4.788, p$<$0.001) ratings revealed statistically significant differences between CushSense and Scuba Fabric. To gain insights into the participant ratings, we analyze the smoothness of the force readings. Smoothness is the reciprocal of the force's second derivative, with the maximum smoothness value calculated for each taxel in every trial round. For CushSense, the smoothness averaged 18.9945 (SD = 10.40469); for Scuba Fabric, it averaged 13.0367 (SD = 7.30359). A paired sample t-test confirmed these differences as statistically significant (p$<$0.05).

To the best of our knowledge, no existing research simultaneously achieves the balance between the spatial resolution (3cm$\times$3cm), passive compliance, and stretchability required to conform to moving joints. While our findings are promising, several avenues warrant further exploration. One interesting topic is the development of automated calibration for precise contact localization and force estimation with CushSense. Another natural progression for future research would be to explore how tactile feedback provided by CushSense can be integrated into closed-loop control policies to enhance pHRI further.

\section{Acknowledgment}
This work was partly funded by National Science Foundation IIS \#2132846, CAREER \#2238792, and DARPA under Contract HR001120C0107. We thank Aparajito Saha for his help with the initial circuit design.

\newpage






\bibliographystyle{ieeetr}
\bibliography{main}

\end{document}